\documentclass[sigconf, screen]{acmart}
\usepackage{multirow}
\usepackage{enumitem}
\AtBeginDocument{%
  }

\acmISBN{978-1-4503-XXXX-X/2018/06}

\renewcommand\footnotetextcopyrightpermission[1]{}

\acmSubmissionID{2025-2301}

\begin{document}

\title{Unified Medical Image Segmentation with \\ State Space Modeling Snake}

\author{Ruicheng Zhang}
\affiliation{%
  \institution{Sun Yat-sen University}
  \city{Shenzhen}
  \state{Guangdong}
  \country{China}
}

\author{Haowei Guo}
\affiliation{%
\institution{Sun Yat-sen University}
  \city{Shenzhen}
  \state{Guangdong}
  \country{China}
  }

\author{Kanghui Tian}
\affiliation{%
\institution{Sun Yat-sen University}
  \city{Shenzhen}
  \state{Guangdong}
  \country{China}
  }

\author{Jun Zhou}
\affiliation{%
\institution{Tsinghua Shenzhen International Graduate School}
  \city{Shenzhen}
  \state{Guangdong}
  \country{China}
  }

\author{Mingliang Yan}
\affiliation{%
\institution{Beijing University of Posts and Telecommunications}
  \city{Beijing}
  \country{China}
  }

\author{Zeyu Zhang}
\affiliation{%
\institution{The Australian National University}
  \city{Canberra}
  \country{Australia}
  }

\author{Shen Zhao}
\affiliation{%
\institution{Sun Yat-sen University}
  \city{Shenzhen}
  \state{Guangdong}
  \country{China}
  }
\authornote{Corresponding author.}



\renewcommand{\shortauthors}{Zhang et al.}

\begin{abstract}
Unified Medical Image Segmentation (UMIS) is critical for comprehensive anatomical assessment but faces challenges due to multi-scale structural heterogeneity. Conventional pixel-based approaches, lacking object-level anatomical insight and inter-organ relational modeling, struggle with morphological complexity and feature conflicts, limiting their efficacy in UMIS. We propose Mamba Snake, a novel deep snake framework enhanced by state space modeling for UMIS. Mamba Snake frames multi-contour evolution as a hierarchical state space atlas, effectively modeling macroscopic inter-organ topological relationships and microscopic contour refinements. We introduce a snake-specific vision state space module, the Mamba Evolution Block (MEB), which leverages effective spatiotemporal information aggregation for adaptive refinement of complex morphologies. Energy map shape priors further ensures robust long-range contour evolution in heterogeneous data. Additionally, a dual-classification synergy mechanism is incorporated to concurrently optimize detection and segmentation, mitigating under-segmentation of microstructures in UMIS. Extensive evaluations across five clinical datasets reveal Mamba Snake’s superior performance, with an average Dice improvement of 3\% over state-of-the-art methods.
\end{abstract}

\vspace{-2em}



\begin{teaserfigure}
  \setlength{\abovecaptionskip}{-0.01em}   
  \includegraphics[width=\textwidth]{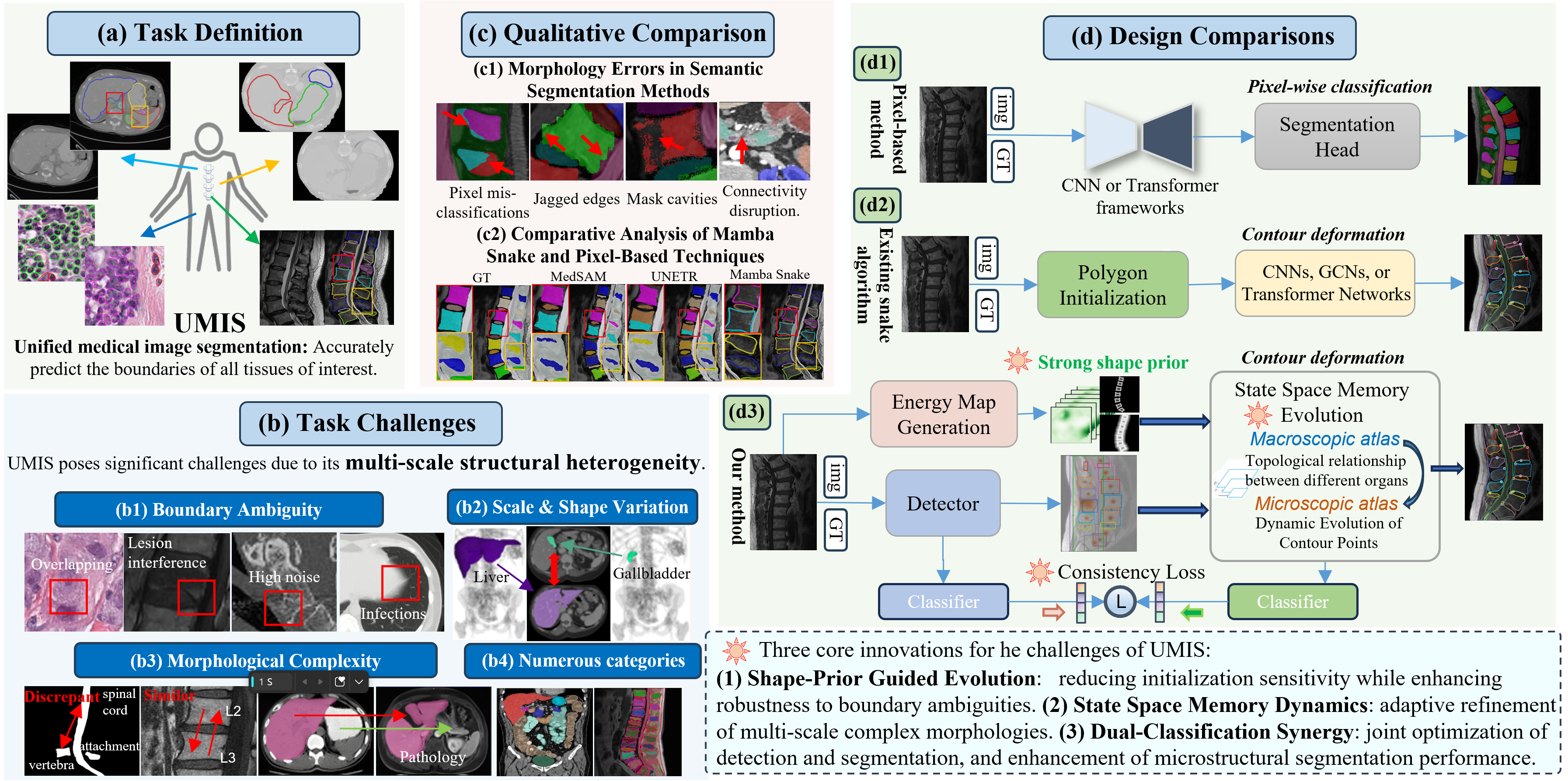}
  \caption{(a) Task definition on Unified Medical Image Segmentation (UMIS). (b) Task challenges of UMIS. (c) Our model's motivation in qualitative view. (d) The design comparison between the pixel-based segmentation methods, the existing snake models, and our Mamba Snake.}
  \label{fig:f1}
\end{teaserfigure}

\maketitle

\section{Introduction}
\label{sec:intro}

Unified Medical Image Segmentation (UMIS) is defined as an advanced framework designed to delineate the boundaries of all regions of interest within a medical image. ``Unified" highlights its ability to accurately segment tissues irrespective of their number, shape, size, or imaging modality (Fig.\ref{fig:f1}(a)), which is essential for clinical comprehensive assessments. For example, UMIS is important in cancer diagnosis and treatment planning because tumor lesions often involve multiple tissues \cite{2022Multi_organ}. Additionally, UMIS is also critical in radiotherapy dose delivery, as it can guide radiotherapy to deliver sufficient doses to affected tissues while minimizing the risk of damage to critical organs \cite{2022head_neck}. 


\begin{figure*}[!t]
\centering
\setlength{\abovecaptionskip}{-0.02em}   
\includegraphics[width=0.9\textwidth]{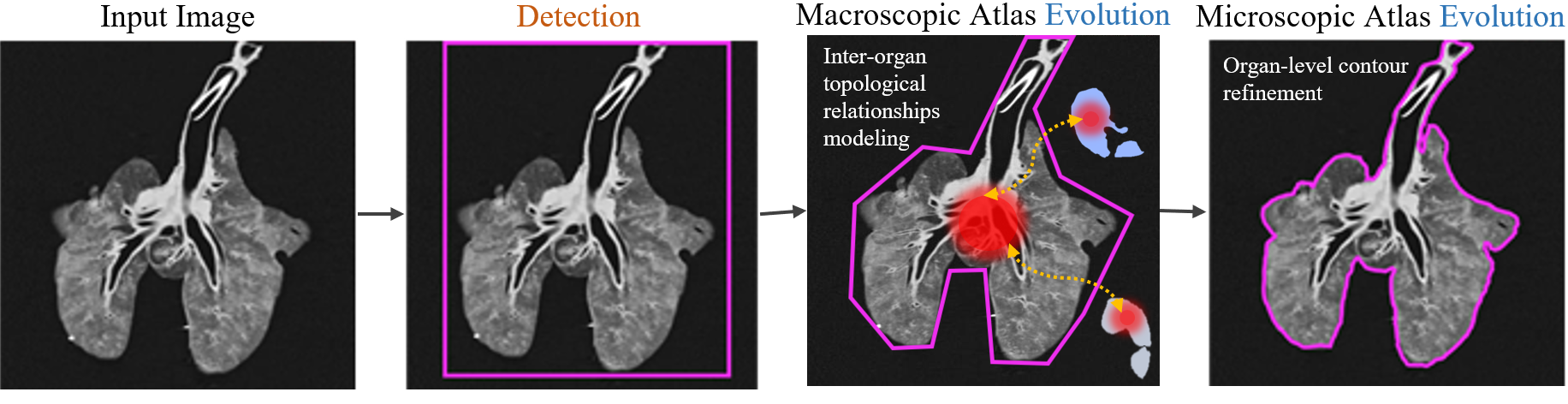}
\caption{ Illustration of the Mamba Snake segmentation pipeline.
}
\label{fig:pip}  
\vspace{-0.2in}
\end{figure*}

While single-target medical image analysis has achieved remarkable success, UMIS remains particularly challenging due to the inherent \textbf{multi-scale structural heterogeneity} across anatomical targets. First, adverse imaging conditions often result in numerous blurred boundaries within medical images; the close proximity and overlap of organs can further intensify this boundary ambiguity (Fig. \ref{fig:f1}(b1)). Second, anatomical  structures in UMIS exhibit nested morphological variations across spatial scales, involving shape, size, position, and orientation (Fig.\ref{fig:f1}(b2)). At the organ level, macroscopic shape differences between vertebrae and intervertebral discs make accurate modeling of their boundaries particularly difficult. At the sub-organ level, adjacent vertebrae display microscopical texture similarity while maintaining similar macro-shapes, posing significant challenges for distinguishing them. Pathological deformations may distort global organ geometry through local lesion morphology (Fig.\ref{fig:f1}(b3)). Furthermore, in the context of UMIS, the organ features across a wide array of categories exhibit significant variability (Fig.\ref{fig:f1}(b4)), which complicates comprehensive feature learning and network optimization. The proportions of organ features at different scales are extremely uneven. Large-scale organs predominantly influence low-frequency feature responses, whereas smaller structures rely on high-frequency details. This spectral disparity, combined with feature interference, results in the under-segmentation of fine structures \cite{2022Multi_organ, 2024medsem}.

Existing methods predominantly employ pixel-based architectures for single-target segmentation across various modalities \cite{2021cycoseg, 2021medical_transformer, 2023vertebrae, 20223d_vessel, 2025UM-net}, whereas efforts to explore Universal Medical Image Segmentation (UMIS) remain limited. Current approaches to UMIS primarily adopt two paradigms: simultaneous prediction frameworks \cite{2021medical_transformer, 2022head_neck, 2022Multi_organ} and medical-adapted Segment Anything Models \cite{2023sa, 2024medsem}. Despite their notable progress, these pixel-wise prediction methods exhibit two significant limitations when addressing UMIS: \textbf{\emph{(1) Inadequate Modeling of Structural Relationships:}} Due to a lack of holistic perception at the object level, current architectures struggle to effectively capture inter-organ contextual relationships in UMIS, such as anatomical hierarchies, spatial arrangements, and topological dependencies. This shortfall frequently leads to disrupted connectivity or pixel misclassification (see Fig. \ref{fig:f1}(c1)). For example, the continuity of the small intestine may be inaccurately segmented, or the natural boundaries between lung lobes may be disregarded, yielding segmentation results that diverge from biological principles \cite{2024medsem}. \textbf{\emph{(2) Sensitivity to Morphological Variations:}} The prevailing pixel-based framework demonstrates instability, as it is susceptible to interference from singular morphological structures in UMIS, such as pathological deformations, resulting in mask cavities and jagged edges (see Fig. \ref{fig:f1}(c1)). These challenges arise from the inherent absence of topological constraints and a comprehensive understanding of organ structures at the object level within pixel-wise frameworks, leading to limited robustness against multi-scale structural heterogeneity.

Deep snake algorithms, integrating the classical active contour model \cite{snake} with deep learning techniques, present a promising alternative for UMIS. In contrast to conventional pixel-based approaches, deep snake algorithms focus on object-level contour prediction through a progressive refinement workflow: \textbf{\textit{generating initial contours}} followed by \textbf{\textit{iterative contours evolution}}. This methodology offers two key advantages for UMIS. First, the coarse-to-fine workflow explicitly facilitates structural relationship modeling by decoupling coarse organ-level shape prediction from subsequent boundary refinements. This design preserves anatomical size hierarchies and spatial interdependencies while mitigating feature interference in densely packed organ scenarios of UMIS \cite{2024medsem}. Second, contour deformation, guided by morphological constraints and point interactions, yields topologically consistent boundaries, significantly reducing the occurrence of abnormal morphological outcomes in UMIS. Despite these strengths, current snake-based methods face limitations in UMIS applications. First, errors in the detection phase can propagate and accumulate, adversely affecting subsequent segmentation accuracy. Second, suboptimal detection boxes may lead to evolutionary stagnation or aberrant convergence. Third, existing snake frameworks \cite{snake, deep_snake, data_MRAVBCE} often overlook dynamic characteristics and historical information of boundary deformation, thereby constraining their evolution capacity and frequently resulting in overly smoothed contours.


Recently, state space models (SSMs), such as Mamba \cite{mamba}, have attracted significant interest from researchers for their ability to provide a global receptive field with linear complexity relative to sequence length \cite{mamba2}. The spatiotemporal dynamics of contour point evolution align well with the properties of state space transitions, inspiring us to employ visual SSMs to drive this process. However, existing SSMs exhibit inherent causal properties, impeding isotropic aggregation of surrounding point information for contour points. Additionally, in visual tasks, these models prioritize spatial receptive fields across patches, often neglecting the temporal characteristics essential to iterative evolution.

To overcome these limitations, we propose Mamba Snake, an advanced deep snake framework augmented with a customized vision mamba block for UMIS, as shown in Figure \ref{fig:f1}(d3). Mamba Snake innovatively models the multi-contour evolution process as hierarchical dynamic state space atlas, where the macroscopic atlas captures the topological relationship among different organs, and the microscopic atlas focuses on the contour evolution of individual organs. Within this framework, we further propose three core innovations to tackle the challenges of UMIS:
\textbf{1. Shape-Prior Guided Evolution:} A boundary distance transform energy map provides continuous anatomical guidance across scales, reducing initialization sensitivity and enhancing robustness to boundary ambiguities. 
\textbf{2. State Space Memory Dynamics:} A pyramid evolution scheme with spatiotemporal memory mechanism models contour deformation as discrete state transitions, facilitating adaptive refinement of complex multi-scale morphologies. We designed a snake-specific visual state space module, the Mamba Evolution Block (MEB), which uses circular convolution to aggregate contour point spatial information and captures dynamic evolution features through temporal hidden states.
\textbf{3. Dual-Classification Synergy:} A consistency-constrained multi-task architecture jointly optimizes detection and segmentation via dual classification heads. By introducing supplementary soft supervision that prioritizes microstructures, it effectively mitigates microstructural under-segmentation.
Comprehensive validation across five clinical domains confirms Mamba Snake's superior performance with 3\% average Dice improvement over current best methods.

The main contributions of Mamba Snake are as follows.
\vspace{-0.25cm}
\begin{enumerate}
\item We propose a novel deep snake framework Mamba Snake that integrates state space modeling for UMIS. Mamba Snake establishes a hierarchical state space atlas with macroscopic inter-organ topological modeling and microscopic contour evolution tracking. 

\item We propose a contour evolution paradigm that integrates energy shape-prior  guidance with state space memory dynamics, enabling complex boundary refinement through continuous state transitions while maintaining multi-organ topological coherence.

\item We design a new snake-specific state space module for effectively driving contour evolution, which utilizes circular convolution to capture spatial dependencies without causal constraints and retention of historical hidden state to enhance temporal modeling.


\item We implement a dual-classification mechanism that synchronizes detection and segmentation through multi-task supervision, effectively mitigating error propagation in multi-organ scenarios by 47\%.

\end{enumerate}

\begin{figure*}[!ht]
\centering
\setlength{\abovecaptionskip}{-0.03em}   
\includegraphics[width=1.0 \textwidth]{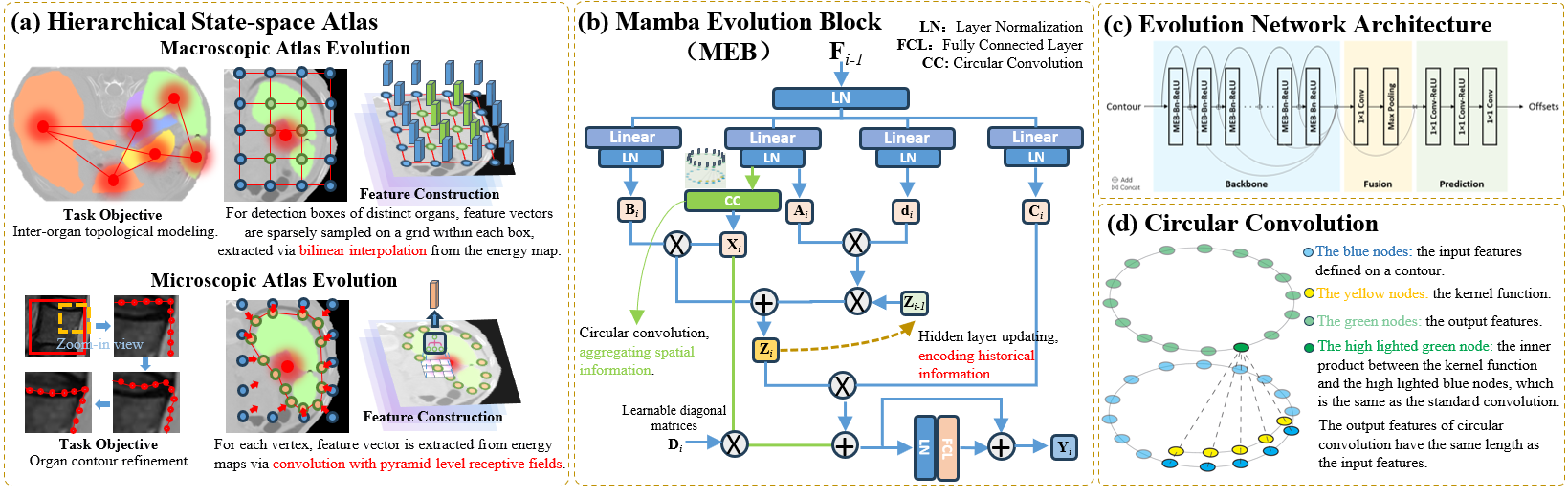}
\caption{ (a) Schematic of the hierarchical state space atlas. (b) Illustration of the Mamba Evolution Block (MEB). (c) Architecture of the contour evolution network. (d) Illustration of the circular convolution principle.
}
\label{fig2}  %
\vspace{-0.5em}
\end{figure*}

\vspace{-1em}
\section{Related Work}
\label{sec:relatedwork}

\subsection{Unified Medical Image Segmentation}

Unified medical image segmentation, which precisely segments all regions of interest across diverse imaging modalities, remains an essential yet challenging task due to multi-scale structural heterogeneity. 

Recent advancements in UMIS have focused on enhancing model adaptability and generalization (\cite{2022Multi_organ, 2024medsem, ESP-MedSAM, CLIP-Driven, compare-sammed2d, zhang2025gamed}). Works such as \cite{ESP-MedSAM, 2024medsem, compare-sammed2d} adapt the Segment Anything Model (SAM) (\cite{sam}) to the medical domain through extensive data training \cite{crisp-sam_huyin}. \cite{Continual_Segment} introduces a continual semantic segmentation framework to dynamically incorporate new classes. \cite{CLIP-Driven} leverages label text embeddings to support multi-class segmentation. \cite{2022head_neck} proposes a stratified segmentation approach to address structures of varying complexity. These models exhibit strong generalization and zero-shot segmentation capabilities; however, their accuracy remains limited in resolving boundary ambiguity, morphological variability, and microstructures. Moreover, modern UMIS models trend toward large parameter counts and complex architectures, complicating their deployment in practical clinical settings.

\vspace{-1em}
\subsection{Deep Snake Model}

Deep snake models \cite{deep_snake,PolarMask,PolyTransform} extend traditional active contour models \cite{snake} by integrating deep learning techniques. These methods represent object shapes as sequences of contour points, regressing the point coordinates in a data-driven approach. \cite{PolarMask} models instance contours in polar coordinates, reformulating instance segmentation as dense distance regression.  \cite{deep_snake}, \cite{GCN}, and \cite{Polygonalformer} adopt the similar snake algorithm pipeline, utilizing CNNs, GCNs, and Transformers, respectively, to predict point-wise offsets for contour evolution.

Compared to conventional pixel-based approaches, deep snake models can robustly generate smooth and accurate object-level anatomical contours across diverse organs under complex conditions. Despite the advantages, their application to challenging UMIS tasks remains unexplored. In multi-scale structural heterogeneity, issues such as blurred boundaries, spurious edges, image noise, significant morphological variations, and unpredictable lesions exacerbate problems like initialization box misalignment, over-smoothed contours, and missed segmentation of small structures. Addressing these challenges requires more advanced evolution strategies and more robust prior guidance.


\subsection{State Space Model}
Recently, state space models, such as Mamba \cite{mamba}, have attracted significant interest from researchers in both language \cite{mamba, mamba2} and vision tasks \cite{VM, VMamba, mamba_v2, mim_huyin}. The S6 block \cite{mamba}, in particular, provides a global receptive field and demonstrates linear complexity relative to sequence length, offering an efficient alternative for snake models. However, its inherent causal properties and limited consideration of temporal evolution characteristics render it unsuitable for direct application to snake models.

\section{Methodology}

\subsection{Overview}

Mamba Snake introduces a novel deep snake framework underpinned by the Mamba Evolution Block (MEB), designed for UMIS to address multi-scale structural heterogeneity. The framework unfolds in two key phases, as illustrated in Fig.\ref{fig:pip}: \textbf{(1) Detection Stage:} A detector generates initial contours by predicting bounding boxes for target tissues. \textbf{(2) Evolution Stage:} These contours are represented as a hierarchical state space atlas, with the MEB facilitating the tracking of subtle deformations to achieve precise boundary delineation.

\subsection{Shape-Prior Guided Snake Evolution}
Incorporating shape prior knowledge into segmentation algorithms has proven useful for obtaining more accurate and plausible results \cite{shape-constraint}. Traditional snake algorithms rely on low-level image features such as grayscale gradients to guide contour evolution. However, this weak guidance proves insufficient for handling the complexities of multi-organ medical images, which feature intricate backgrounds, blurred boundaries, and diverse contour morphologies \cite{zhang2025gamed}.  
 
In Mamba Snake, we design the Energy Shape Prior Map (ESPM) to regulate contour point coordinates, enhancing robustness against complex image characteristics caused by multi-scale structural heterogeneity. This approach is also proven to be beneficial for obtaining more plausible results and avoiding unreasonable morphology errors  \cite{shape-constraint}.

The ESPM establishes continuous boundary attraction fields through learnable energy mapping. The pixel-level energy values $E(x,y)$ is constructed using a boundary distance transform:
\begin{equation}
E(x,y) = \mathcal{D}_T(I) \ast G_\sigma + \lambda \|\nabla I \|^{-0.5},
\end{equation}
\noindent where $\mathcal{D}_T(I)$ represents the distance transform from predicted tissue boundaries to coordinates $I(x,y)$, $G_\sigma$ denotes Gaussian smoothing, and the edge potential term $\|\nabla I\|^{-0.5}$ amplifies gradient response at weak boundaries. The ESPM provides long-range guidance for contour evolution by enhancing boundary features, thereby reducing sensitivity to initial contour placement and bolstering robustness against boundary ambiguities. The distance transform $\mathcal{D}_T$ intensifies as the proximity to the target boundary decreases, forming attraction basins along these boundaries. We evaluate different distance transform functions, including linear, exponential, and logarithmic functions:
\vspace{-0.5em}
\begin{align}
    \text{Lin:} \quad & {D}_T(I) = 255 - \text{Norm}(D_{I}), \\
    \text{Exp:} \quad & {D}_T(I) = 255 \ast  e^{-\lambda \text{Norm}(D_{I})}, \\
    \text{Log:} \quad & {D}_T(I) = 255 \ast (1-\log(1 + \alpha  \text{Norm}(D_{I}))).
\end{align}
\vspace{-1.5em}

\begin{figure}[!tbh]
    \centering
    \setlength{\abovecaptionskip}{-0.02em}
    \includegraphics[width=1\linewidth]{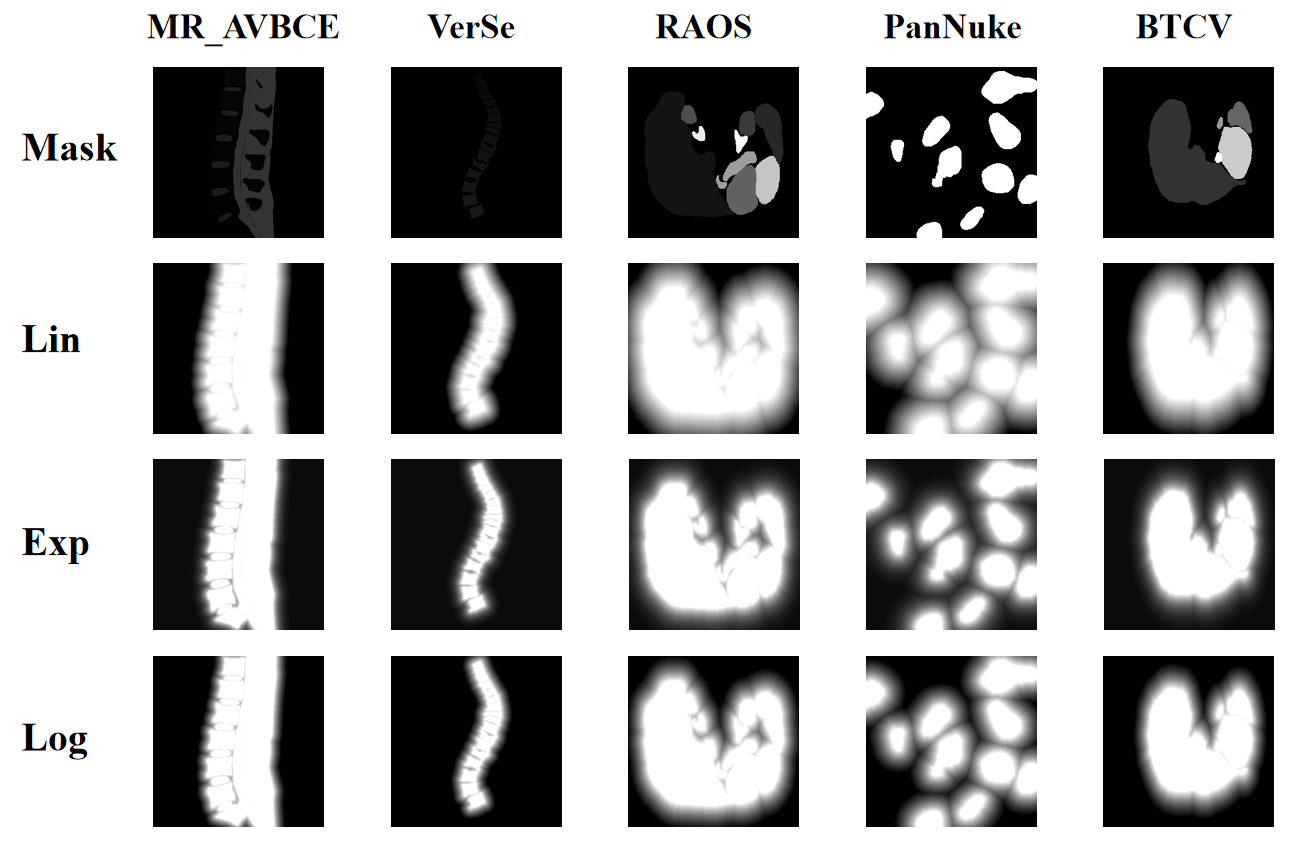}
    \caption{Example energy maps generated from the mask images of five distinct datasets, illustrating how energy values vary based on shape, size, and spatial relationships within each dataset.}
    \label{fig:energy_maps}
    \vspace{-2em}
\end{figure}

Here, $\lambda$ and $\alpha$ are used to control the energy decay rate. Figure \ref{fig:energy_maps} presents example energy maps derived from the mask images of five different datasets, demonstrating the variation in energy distribution across different anatomical structures. Despite variations in energy map morphology, they consistently demonstrate robust performance, as shape priors inherently capture general organ geometry.

\subsection{State Space Memory Dynamics}
Effectively driving contour points to evolve toward target boundaries poses a complex challenge. Traditional snake algorithms utilize empirical attraction functions based on low-level image features, often leading to convergence on local optima. Existing deep snake methods, such as Deep Snake \cite{deep_snake} and PolyTransform \cite{PolyTransform}, typically treat contour evolution as a topological problem, employing CNNs or transformers to directly predict positional offsets. While these approaches are straightforward and effective, they overlook the dynamic and temporal aspects of evolution and fail to capture interdependencies among organ contours, leading to issues like over-smoothing of complex organ shapes and contour overlap in UMIS.

Mamba Snake models multi-contour evolution as a hierarchical state-space atlas encompassing macroscopic and microscopic perspectives. It implements the Mamba Evolution Block, a tailored visual state space module to facilitate evolution drive with spatiotemporal memory, offering a novel solution for contour refinement in UMIS contexts.

\noindent \textbf{Macroscopic Atlas Evolution}

The macroscopic atlas evolution is to model contextual relationships among organs, encompassing anatomical hierarchies, spatial configurations, and topological dependencies, initiating the generation of initial polygons from detection boxes. For a set of \( K \) detection boxes, denoted \( b_k \) (\( k = 1, 2, \dots, K \)) corresponding to distinct organs, feature vectors are sparsely sampled on an \( M \times M \) grid within each box. Since fine-grained boundary delineation is not required at this stage, grid sampling sufficiently characterizes the organs' coarse topological features. The position of the \( m \)-th grid point is represented as \( \mathbf{v}_{k,m} = (v_{k,m,x}, v_{k,m,y}) \in \mathbb{R}^{2} \), with its corresponding feature vector extracted via bilinear interpolation from a feature map \( F \in \mathbb{R}^{128 \times 128 \times 64} \):
\begin{equation}
\mathbf{f}_{k,m} = F(\mathbf{v}_{k,m}), \quad \forall m \in \{1, 2, \dots, M^2\}.
\end{equation}

To integrate spatial context, these vectors are concatenated with their respective coordinates \( \mathbf{v}_{k,j} \), forming the resultant feature vector for the \( k \)-th box:
\begin{equation}
\mathbf{f}_k = \left[ (\mathbf{f}_{k,1}, \mathbf{v}_{k,1}), (\mathbf{f}_{k,2}, \mathbf{v}_{k,2}), \dots, (\mathbf{f}_{k,M}, \mathbf{v}_{k,M^2}) \right].
\end{equation}

The feature vectors $\{ \mathbf{f}_k \}_{k = 1}^K \in \mathbb{R}^{66}$ are then processed by a state-space deformation model to predict positional offsets for 40 edge points per box, as shown in Fig.\ref{fig2}.

\noindent \textbf{Microscopic Atlas Evolution}

The microscopic atlas evolution refines the initial polygon to precisely align with the target organ’s boundary. For a contour with \( N = 128 \) vertices \( \{ \mathbf{x}_i \mid i = 1, 2, \dots, N \} \), a feature vector is constructed for each vertex. The input feature vector \( \mathbf{f}_i \) integrates image-derived features and normalized spatial coordinates:
\begin{equation}
  \quad \mathbf{f}_i = \left[ F(\mathbf{x}_i); \mathbf{x}_i' \right] \in \mathbb{R}^{66},
\end{equation}
where \(F(\mathbf{x}_i)\in\mathbb{R}^{64}\) is extracted from the energy maps \(F \in \mathbb{R}^{128 \times 128 \times 64}\) via a convolution module of pyramid-level receptive fields and \( \mathbf{x}_i'\in \mathbb{R}^{2} \) denotes relative vertex coordinates. Since the deformation should not be affected by the translation of the contour in the image, we subtract \( \mathbf{x}_i' \) by the center coordinates of each detection box.

Using the feature set \( \{ \mathbf{f}_i \}_{i=1}^N \), the microscopic atlas employs the state-space deformation model \( \Psi \) to predict offsets \( \Delta \mathbf{x}_i=\Psi(\mathbf{f}_i) \) and update vertex positions:
\begin{equation}
\mathbf{x}_i' = \mathbf{x}_i + \Delta \mathbf{x}_i, \quad \forall i \in \{1, 2, \dots, 128\}.
\end{equation}

Given the challenge of regressing contour points to the target boundary in a single step, especially when the initial contour points are distant from the target boundary, an iterative approach is adopted. The specific number of iterations is detailed in Ablation Studies.


\noindent \textbf{Mamba Evolution Block}

Previous visual SSMs \cite{VM,VMamba} typically flatten 2D feature maps into 1D sequences using various scanning strategies and process them with the S6 block \cite{mamba_v2}. This approach disrupts the intrinsic structural relationships among feature vectors, limiting contour points in evolving scenarios to only accessing preceding point features, thus failing to integrate information from subsequent points. Moreover, these visual SSMs, prioritize spatial receptive fields, often overlooking the temporal characteristics of iterative contour point motion.

To address these limitations, we propose a novel snake-specific VSSM, the Mamba Evolution Block (MEB). The MEB redefines the state space transition matrix $A$
 as a scalar while expanding the state space dimension, transforming it into a non-causal framework \cite{mamba2}. Specifically, instead of using $A$
 to regulate the retention of hidden states, we leverage it to control the contribution of the current contour point token to the hidden states. Given that contour points are topologically constrained by $N$ neighboring points, the MEB employs circular convolution to aggregate spatial information (see Fig.\ref{fig2}(d)). Additionally, to guide current evolution with historical context, the MEB retains past hidden states to inform present decisions.

At step \( i-1 \), the feature vector of contour point \( k \), denoted \( \mathbf{F}_{i-1,k} \), is mapped via four linear projections into state-space variables at step \( i \): input \( \mathbf{X}_i \), mapping matrices \( \mathbf{B}_i \) and \( \mathbf{C}_i \), transition matrix \( A_i \), and weighting coefficient \( d_i \). Subsequently, \( \mathbf{X}_i \) undergoes circular convolution (see Fig.\ref{fig2}(b)) for adaptive feature interaction, followed by sigmoid activation; \( \mathbf{B}_i \), \( \mathbf{C}_i \), and \( d_i \) are similarly activated and reshaped as state-space parameters.  For clarity, intermediate variables retain their original notation post-transformation. The MEB operates in two phases: first, expanding \( \mathbf{X}_i \) with \( \mathbf{B}_i \) and unrolling scalar recurrences to form a global hidden state \( \mathbf{Z}_i \); second, contracting \( \mathbf{Z}_i \) with \( \mathbf{C}_i \) while skip-connecting the expanded \( \mathbf{X}_i \). The model is expressed as:
\begin{equation}
\mathbf{Z}_i = d_i \mathbf{A}_i (\mathbf{Z}_{i-1} + \mathbf{B}_i \mathbf{X}_i), \ \
\mathbf{Y}_i = \mathbf{C}_i^\top \mathbf{Z}_i + \mathbf{D}_i \mathbf{X}_i,
\end{equation}
where \( \mathbf{D}_i \) is a trainable skip-connection matrix. The latent \( \mathbf{Z}_i \), initialized with \( \mathbf{X}_0 \), encodes state space, incrementally integrating \( \mathbf{B}_i \mathbf{X}_i \) to capture dynamic spatiotemporal features. The adaptive weighting \( d_i \) balances current and historical contributions. By modeling long-term evolution patterns, MEBB excels in iterative contour delineation, particularly for ambiguous boundaries. The historical context embedded in \( \mathbf{Z}_i \) leverages prior cues (e.g., boundary directionality) when current features are unclear, ensuring robust evolution.

\begin{figure*}[!t]
\centering
\setlength{\abovecaptionskip}{-0.02em}   
\includegraphics[width=0.9 \textwidth]{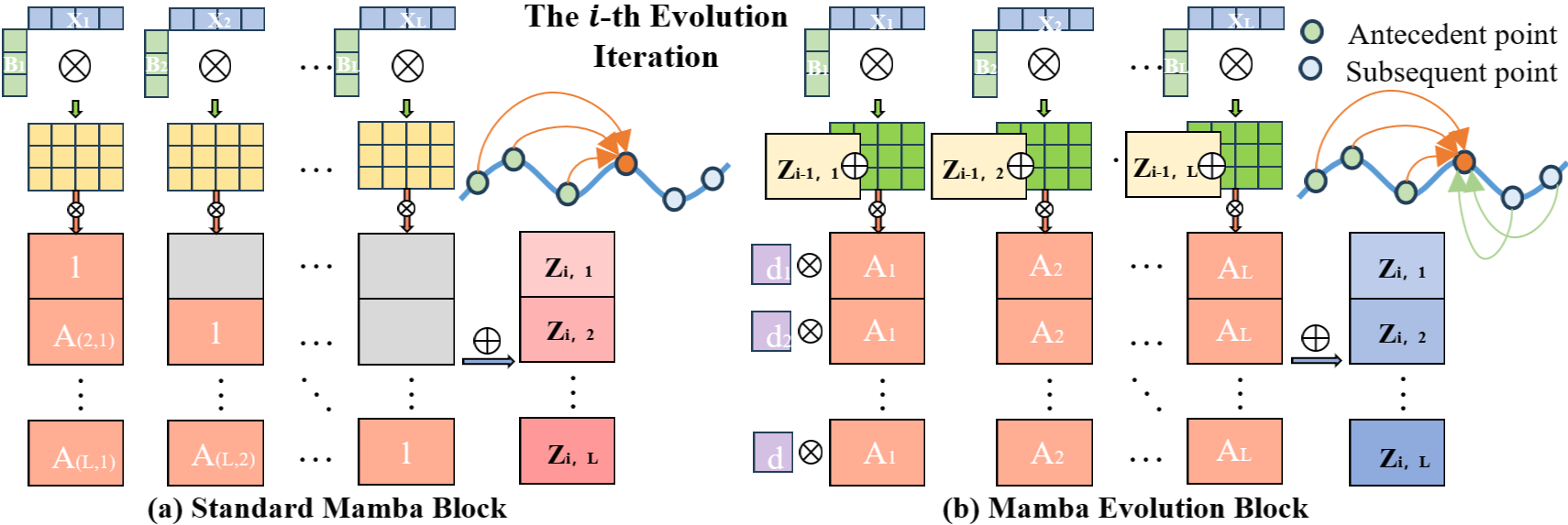}
\caption{ The hidden state generation process for standard mamba and MEB. The standard Mamba restrict the central point to accessing only antecedent points, while MEB allows integrate of features from both antecedent and subsequent points.
}
\label{fig:res}  %
\vspace{-1.5em}
\end{figure*}

\subsection{Dual-Classification Synergy}
Inspired by the success of multi-head classification in self-distillation \cite{Self_Distillatio} and self-supervised learning \cite{Liu2025DualCH}, Mamba Snake introduces Dual-Classification Synergy to concurrently enhance detection and segmentation performance. This approach employs two classification heads: the Detection Classifier \( C_d \) and the Segmentation Classifier \( C_s \). Specifically, \( C_d \) predicts organ category probability vectors \( \mathbf{p}_d \) from multi-scale region proposal features during detection, while \( C_s \) derives probability vectors \( \mathbf{p}_s \) from contour point features during evolution. A weighted average of \( \mathbf{p}_d \) and \( \mathbf{p}_s \) is processed via softmax and assessed against ground-truth labels using cross-entropy loss \( L_H \). Additionally, a consistency loss \( L_S \) enforces alignment between the soft labels of \( \mathbf{p}_d \) and \( \mathbf{p}_s \):
\vspace{-0.5em}
\begin{equation}
\begin{aligned}
L_H &= -\sum_{c=1}^C y_c \log \left( \text{softmax} \left( w_d \mathbf{p}_d + w_s \mathbf{p}_s \right)_c \right), \\
L_S &= K \left( -\sum_{c=1}^C \text{softmax}(\mathbf{p}_d)_c \log \left( \text{softmax}(\mathbf{p}_s)_c \right) \right),
\end{aligned}
\end{equation}
where \( y_c \) is the ground-truth label for class \( c \), \( C \) is the number of classes, \( w_d \) and \( w_s \) are weights (\( w_d + w_s = 1 \)), and \( K \) is a size penalty factor inversely proportional to the ground-truth mask's pixel size. This formulation enhances detection of small organs, addressing under-detection and under-segmentation challenges in UMIS.

The dual-classification strategy leverages target boundary features to refine the detector the detector's edge-learning capability. This results in higher category confidence and tighter detection boxes, which in turn supports more precise contour evolution toward target boundaries, yielding superior segmentation outcomes.

\vspace{-0.2em}
\subsection{Implementation Details}
\noindent \textbf{The Energy Shape Prior Map generation}  \ The  energy map generation network is built upon the EfficientNetV2-S \cite{EfficientNetV2} backbone, followed by deconvolution layers for outputting predictions. Specific network details can be found in the supplementary material. 

\noindent \textbf{Detector} \ We adopt CenterNet \cite{centernet} as the detector for Mamba Snake, which generates class-specific detection boxes to initialize polygonal  contours. CenterNet reformulates the detection task as a keypoint detection problem and achieves an impressive trade-off between speed and accuracy. It is worth noting that Mamba Snake only needs the detection boxes provided by the detector for initializing the polygonal contours. Therefore, the detector can be replaced by any other detection model, such as the Yolo series \cite{YOLOv8}.

\noindent \textbf{Contour evolution} \ We uniformly sample $N$ points from both the ground truth boundary and the snake contour and pair them by minimizing the distance between corresponding poins. Mamba Snake takes the contour features as input and outputs $N$ offsets that point from each vertex to the target boundary point. We set $N$ to 128 in all experiments, which is sufficient to cover most organ shapes. The number of evolutionary iterations is set to 3.

\noindent \textbf{Training Strategy and Loss Functions} \quad
To ensure robust performance, we first pretrain the energy map generation network for accurate distance energy map predictions, followed by joint optimization of the detection and snake evolution processes. In the pretraining phase, the energy map is optimized using the Charbonnier loss, defined as:
\begin{equation}
\mathcal{L}_E = \sqrt{\left\| f_E(P(x, y)) - E_P^{GT} \right\|^2 + \epsilon^2}, \quad \epsilon = 10^{-3},
\end{equation}
where \( E_P^{GT} \) is the ground-truth distance energy value, and \( f_E(\cdot) \) denotes the energy map generation network.

The detection and contour evolution components are jointly trained in an end-to-end manner. 
The detection component employs the loss function \( L_{\text{det}} \) from the original detection model \cite{centernet}.  
The contour evolution loss \( L_{\text{evol}} \) is defined as the mean \(\ell_1\) distance between the predicted and ground-truth contour points:
\begin{equation}
L_{\text{evol}} = \frac{1}{N} \sum_{i=1}^{N} \ell_1 (\tilde{\mathbf{x}}_{i} - \mathbf{x}_{i}^{\text{gt}}), \quad \text{where } N = 128.
\label{eq:evolution_loss}
\end{equation}
where \( \tilde{\mathbf{x}}_{i} \) represents the predicted coordinates of the \(i\)-th contour point, and \( \mathbf{x}_{i}^{\text{gt}} \) is its corresponding ground-truth coordinate.

The total loss of the model is formulated as:
\begin{equation}
L = L_{ex} + L_{ev} + 0.5 L_H + 0.5 L_S + L_{Detector},
\end{equation}
where \( L_H \) and \( L_S \) are the classification losses from the Dual-Classification Synergy, and the weighting coefficients normalize the contributions of each term to a consistent scale. 

\vspace{-0.2em}
\section{Experiments and results}

\begin{figure*}[!ht]
\centering
\setlength{\abovecaptionskip}{-0.02em}   
\includegraphics[width=1.0 \textwidth]{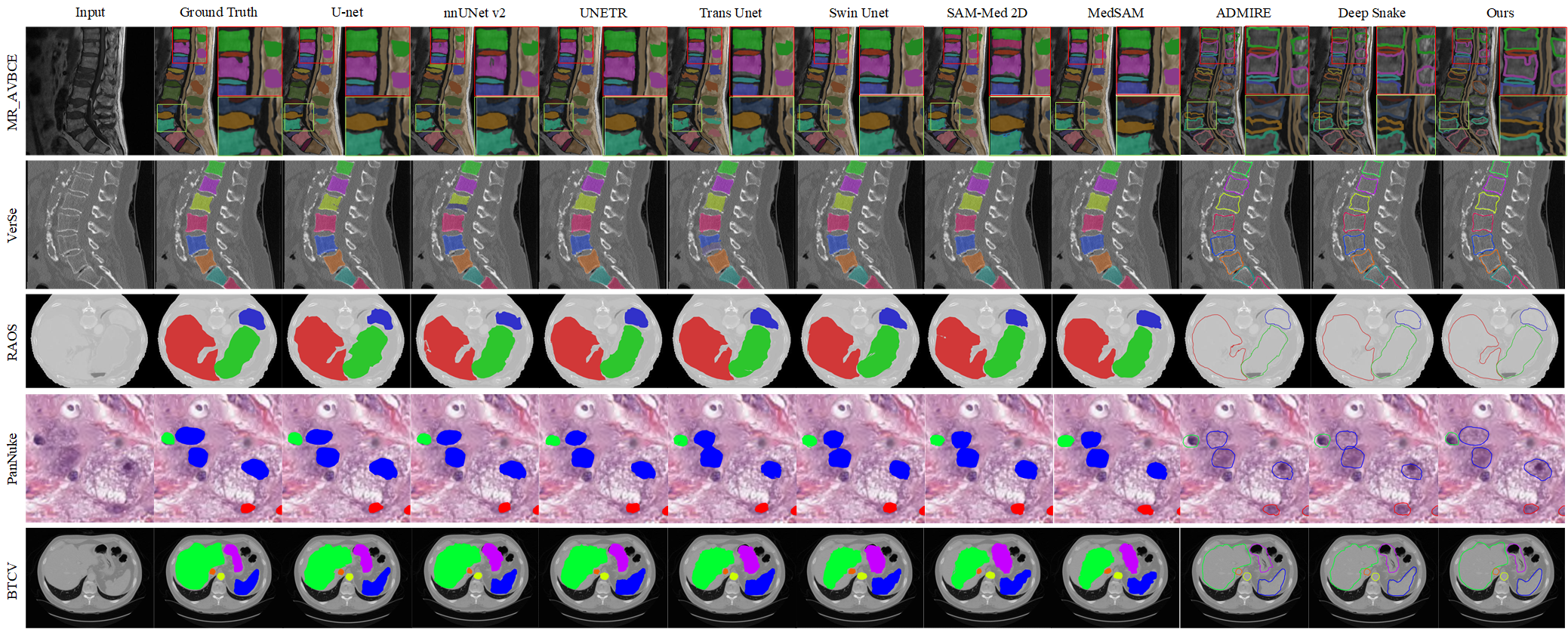}
\caption{ Qualitative comparison of segmentation results between Mamba Snake and other methods across five datasets.
}
\label{fig:res}  %
\vspace{-0.3em}
\end{figure*}

\begin{table*}[h!]
\centering
\renewcommand{\arraystretch}{1}  
\setlength{\tabcolsep}{4pt}  
\resizebox{\textwidth}{!}{
\begin{tabular}{l|ccc|ccc|ccc|ccc|ccc}
\hline
\multirow{2}{*}{\textbf{Model}} & \multicolumn{3}{c|}{\textbf{MR\_AVBCE}} & \multicolumn{3}{c|}{\textbf{Verse}} & \multicolumn{3}{c|}{\textbf{RAOS}} & \multicolumn{3}{c|}{\textbf{PanNuke}} & \multicolumn{3}{c}{\textbf{BTCV}}\\ \cline{2-16}
 & \textbf{mIoU} & \textbf{mDice} & \textbf{mBoundF}  & \textbf{mIoU} & \textbf{mDice}& \textbf{mBoundF} & \textbf{mIoU}  & \textbf{mDice}& \textbf{mBoundF }  & \textbf{mIoU } & \textbf{mDice} & \textbf{mBoundF}  & \textbf{mIoU} & \textbf{mDice} & \textbf{mBoundF} \\ \hline
U-Net \cite{compare-unet} &  80.02 & 86.32 & 81.33 & 78.25 & 84.29 & 79.24 & 80.52 & 87.55 & 83.36 & 78.26 & 85.75 & 85.82 & 81.04 & 87.02 & 87.29\\
nnUNet V2 \cite{compare-nnunet} & 83.76 & 90.64 & 86.54 & 83.29 & 90.63 & 86.14 & 83.42 & 91.12 & 88.66 & 83.35 & 91.26 & 88.26 & 83.74 & 91.03 & 89.19\\
UNETR \cite{unetr} &  83.34 & 90.35 & 85.74 & 82.46 & 88.81 & 84.39 & 81.69 & 90.26 & 87.86 & 83.52 & 91.56 & 88.73 & 82.86 & 90.43 & 88.75\\
Trans Unet \cite{2021transunet} & 81.52 & 88.42 & 85.24 & 81.95 & 87.34 & 83.49 & 81.97 & 88.16 & 85.12 & 82.14 & 90.83 & 86.29 & 81.92 & 88.54 & 88.19\\
Swin Unet \cite{cao2021swinunet} & 83.16 & 89.52 & 85.34 & 81.59 & 87.36 & 85.44 & 82.49 & 89.96 & 87.79 & 82.29 & 90.97 & 87.57 & 82.06 & 90.03 & 88.95\\
SAM-Med 2D \cite{compare-sammed2d} & 81.26 & 87.46 & 82.49 & 79.29 & 85.90 & 82.77 & 80.79 & 87.62 & 86.65 &80.48 & 87.85 & 85.46 & 81.65 & 87.39 & 87.52 \\
MedSAM \cite{2024medsem} & 83.62 & 89.46 & 86.28 & 80.25 & 86.50 & 83.46 & 81.60 & 87.89 & 84.29 & 81.69 & 89.27 & 86.19 & 82.01 & 89.19 & 88.48 \\ \hline
ADMIRE \cite{data_MRAVBCE} & 81.71 & 88.38 & 85.53 & 79.84 & 86.52 & 85.78 & 81.60 & 87.89 & 87.44 & 82.29 & 90.90 & 88.29 & 82.25 & 89.53 & 87.09 \\
Deep Snake \cite{deep_snake} & 79.49 & 86.41 & 85.03 & 79.02 & 86.04 & 84.78 & 80.53 & 86.95 & 87.07 & 81.74 & 87.43 & 87.19 & 81.94 & 88.03 & 87.69 \\ \hline
\textbf{Ours} & \textbf{85.25} & \textbf{93.76} & \textbf{90.37} & \textbf{84.96} & \textbf{92.31} & \textbf{90.23} & \textbf{85.75} & \textbf{93.92} & \textbf{90.75} & \textbf{85.14} &\textbf{92.35} & \textbf{90.36} & \textbf{86.04} &\textbf{94.27} & \textbf{92.36}\\
\hline    
\end{tabular}
}
\caption{  \textbf{Quantitative results.} All metrics are reported as \% values. Bold values indicate the best results in the table.}
\label{tab:comparison}
\vspace{-0.25in}
\end{table*}

\subsection{Experiment Configurations}
\textbf{Datasets}\quad This study evaluates the performance of the Mamba Snake model using five prominent multi-organ segmentation datasets: MR\_AVBCE (spine, MRI) \cite{data_MRAVBCE}, VerSe (spine, CT) \cite{data_verse}, RAOS (abdomen, CT) \cite{data-RAOS}, BTCV (abdomen, CT) \cite{dataset_btcv}, and PanNuke (cells, microscopy) \cite{data-PanNuke}. These datasets span diverse human tissues, including spine, abdomen, and cells, and exhibit typical characteristics of multi-scale structural heterogeneity, such as numerous categories, significant morphological variations, severe pathological conditions, and blurred boundaries. Details of the datasets are provided in the supplementary materials.

\noindent \textbf{Evaluation Metrics}\quad We employ three established metrics commonly used in medical image segmentation: mean Intersection over Union (mIoU), mean Dice Similarity Coefficient (mDice) \cite{metric-Dice}, and mean Boundary F (mBoundF) \cite{boundf}. These are defined as follows:  
\begin{equation}
\begin{gathered}
\text{mIoU} = \frac{1}{N} \sum_{i=1}^{N} \frac{|A_i \cap B_i|}{|A_i \cup B_i|}, \ \ \
\text{mDice} = \frac{1}{N} \sum_{i=1}^{N} \frac{2 |A_i \cap B_i|}{|A_i| + |B_i|}, \\
\text{mBoundF} = \frac{1}{5} \sum_{n=1}^{5} \text{mDice}_n (\partial A_n, \partial B_n),
\end{gathered}
\end{equation}
where \(A_i\) denotes the ground truth for the \(i\)-th organ, \(B_i\) represents the predicted segmentation, and \(\partial A_n\) and \(\partial B_n\) denote the boundaries of the ground truth and predicted masks\footnote{Due to discontinuous boundaries in some pixel-based segmentation methods, we replaced the overlap calculation between GT and predicted boundaries in mBoundF with the overlap between the GT boundary and the nearest points on the predicted mask. This approach provides a more lenient evaluation metric.}, respectively, with n indicating boundary width (1 to 5 pixels). These metrics ensure a comprehensive assessment of both the model's segmentation accuracy (mIoU and mDice) and boundary-delineation quality (mBoundF).

\noindent \textbf{Implementation Details}\quad The model is implemented in Python 3.7 with PyTorch 1.9.0, and all experiments are conducted on two NVIDIA RTX 3090 GPUs, each with 24GB of memory. We use the AdamW optimizer with a batch size of 8 and an initial learning rate of \(1 \times 10^{-4}\), which is gradually decreased to \(1 \times 10^{-6}\) using a cosine annealing strategy to facilitate model convergence.

\vspace{-0.4em}
\subsection{Comparing Experiments}
We compare Mamba Snake with two categories of state-of-the-art methods: (1) pixel-based segmentation models, including U-Net \cite{compare-unet}, nnUNet v2 \cite{compare-nnunet}, UNETR \cite{unetr}, TransUNet \cite{chen2021transunet}, SwinUNet \cite{cao2021swinunet}, SAM-Med2D \cite{compare-sammed2d}, and MedSAM \cite{2024medsem}, and (2) contour-based segmentation models, such as ADMIRE \cite{Zhao2023Attractive} and Deep Snake \cite{deep_snake}.

\subsubsection{Quantitative Results}
\textbf{Overall.} Table \ref{tab:comparison} presents the comparative results across the MR\_AVBCE, Verse, RAOS, PanNuke, and BTCV datasets. Our model achieves the best performance on all five datasets in all the three evaluation metrics. Specifically, on the most challenging MR\_AVBCE dataset, our model outperforms the second-best method by 1.67\% in mIoU, 1.68\% in mDice, and a significant 4.09\% in mBoundF. These results not only underscore the overall superiority of our model but also highlight its exceptional capability in boundary precision, as evidenced by the substantial improvement in the mBoundF metric. The significant margin in mBoundF reflects the robustness of the deep snake algorithm in generating smooth and accurate contours, with the integration of the Mamba evolution strategy and energy prior guidance further  enhancing the adaptability of the contour delineation process. Our approach markedly improves the precision of object boundary segmentation, which is of paramount importance in applications such as medical image segmentation.

\noindent \textbf{Considering substructures.} For the MR\_AVBCE dataset, we present the segmentation results of our model compared to other comparison methods across multiple fine-grained substructure categories. These 26 categories are chosen based on their prevalence in the dataset, with three evaluation metric sets calculated and detailed in the supplementary material. Our model outperforms others, achieving the highest mIoU and mDice scores for 12 substructures and the top mBoundF scores for 14 substructures.

\subsubsection{Qualitative Results}
In Fig.\ref{fig:res}, we present segmentation comparisons for selected slices across five datasets, including magnified views. The results indicate that most pixel-based methods falter in differentiating closely spaced and challenging categories—such as tumor-affected vertebrae in MR\_AVBCE, T12/L1 transitions in VerSe, and overlapping boundaries in PanNuke—exhibiting errors including pixel misclassifications, inconsistent segmentations, and mask voids. Conversely, Mamba Snake demonstrates superior performance by accurately delineating object-level category boundaries. Visualization results further reveal that comparative methods fail to predict precise boundaries for small, complex structures—e.g., blurred inter-vertebral discs in MR\_AVBCE, minor vertebrae in VerSe, small organs in BTCV, and dense nuclei clusters in PanNuke—whereas Mamba Snake yields satisfactory outcomes. These findings highlight the effectiveness of the model’s state-space memory evolution in addressing multi-scale structural heterogeneity.

\begin{figure}[!t]
\centering
\setlength{\abovecaptionskip}{-0.02em}   
\includegraphics[width=0.48 \textwidth]{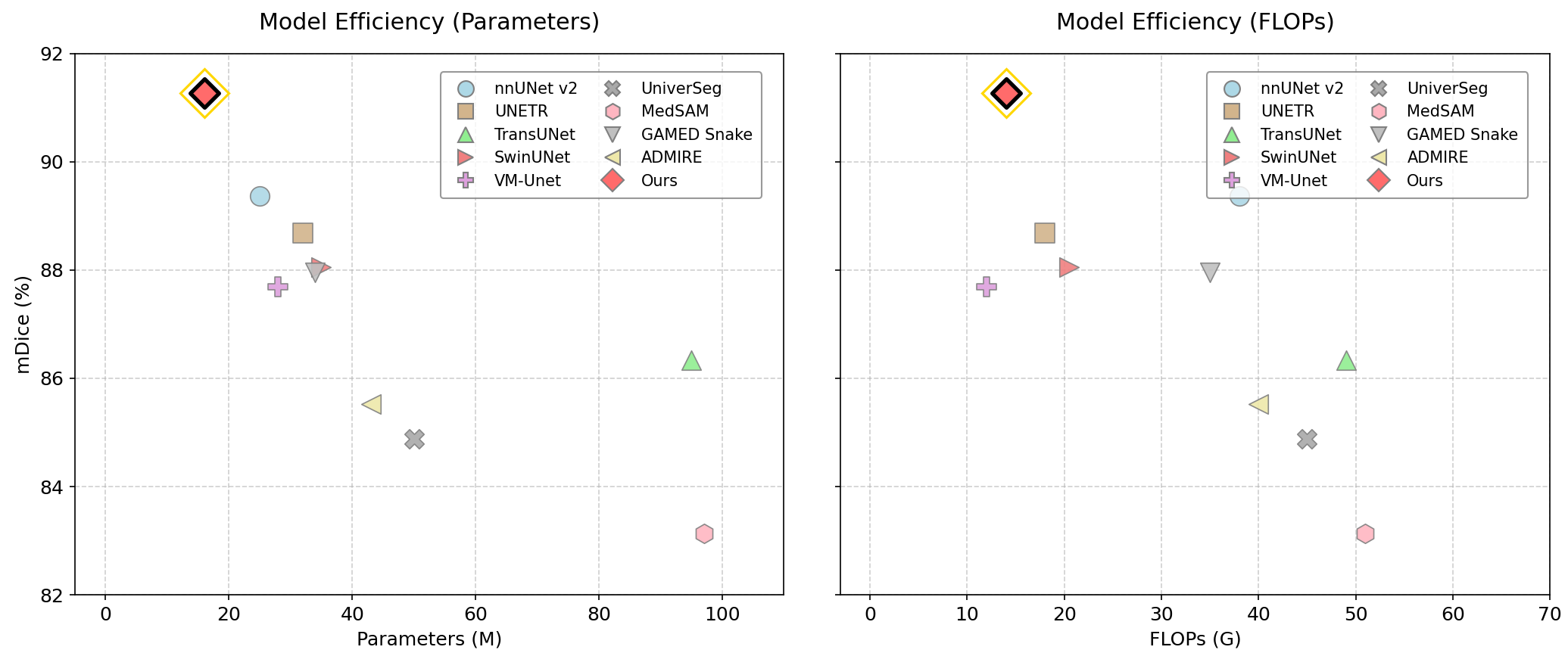}
\caption{ Model efficiency analysis. The computational efficiency and inference speed  of our method and the previous SOTA methods to process a 512 × 512 image.
}
\label{fig:eff}  %
\vspace{-0.2in}
\end{figure}

\subsection{Ablation Study}
\subsubsection{\textbf{Key Components} }
We evaluate the contribution of model’s key components on the MR\_AVBCE dataset as it features the highest number of semantic categories, the most complex organ morphologies, and severe lesion invasions, exemplifying typical UMIS multi-scale structural heterogeneity. The experimental results are presented in Table \ref{tab:as}, revealing the following:

\textit{(1) ESPM Enhances Performance Consistently}. The integration of ESPM yields approximately a 3\% improvement in mDice and mIoU, attributed to robust shape prior guidance that enhances model resilience to blurred boundaries and complex backgrounds. It is worth to note that this strategy also effective for pixel-based segmentation methods.

\textit{(2) SSMD Significantly Improves Accuracy over 4\% on all metrics}. This improvement stems from effective state-space modeling, where historical evolution memory boosts segmentation precision for complex morphologies; the hierarchical atlas design explicitly accounts for inter-organ topological relationships, reducing contour overlaps in dense UMIS scenarios.

\textit{(3) DCS Optimizes Fine Boundary Segmentation}. The Segmentation Classifier leverages contour feedback to refine detector learning of organ boundary features, resulting in greater boundary precision (mBoundF) improvements compared to region overlap metrics (mIoU and mDice). Additionally, targeted supervision of small structures reduces organ under-segmentation by 47\%.

\begin{table}[h!]
\centering
\renewcommand{\arraystretch}{0.92}  
\setlength{\tabcolsep}{5.5pt}  
\resizebox{0.48\textwidth}{!}{
\begin{tabular}{l|ccc}
\hline
\multirow{1}{*}{\textbf{Model}} & \multicolumn{1}{c}{\textbf{mIoU}} & \multicolumn{1}{c}{\textbf{mDice }} & \multicolumn{1}{c}{\textbf{mBoundF}} \\  \hline
Baseline & 79.49 & 87.41 & 85.03\\
+ Energy Shape Prior Map (ESPM) & 82.27 ($3.5\%\uparrow$) & 90.09 ($2.8\%\uparrow$) & 87.11 ($2.4\%\uparrow$) \\
+ State Space Memory Dynamics (SSMD)& 84.01 ($5.7\%\uparrow$) & 91.23 ($4.3\%\uparrow$) & 89.14 ($4.8\%\uparrow$) \\
+ Dual-Classification Synergy (DCS)& 81.79 ($2.9\%\uparrow$) & 89.96 ($2.9\%\uparrow$) & 87.65 ($3.1\%\uparrow$) \\
Mamba Snake & \textbf{85.25} $\textcolor{red}{(7.2\%\uparrow)}$ & \textbf{93.76} $\textcolor{red}{(7.3\%\uparrow)}$ & \textbf{90.37} $\textcolor{red}{(6.3\%\uparrow)}$ \\
\hline
\end{tabular}
}
\caption{ \textbf{Ablation studies on MR\_AVBCE dataset.} All metrics are reported as \% values. The Baseline is a direct combination of Deep Snake \cite{deep_snake} with CenterNet \cite{centernet}. }
\label{tab:as}
\vspace{-0.1in}
\end{table}

\begin{table}[!h]
\centering
\scalebox{0.85} {
\begin{tabular}{l|ccccc}
\textbf{Evolution Iterations} & {1} & {2} & {3} & {4} & {5} \\ \hline
\textbf{mDice}(\%) & 83.54 & 87.17 & \textbf{93.76} & 91.06 & 90.43 \\
\textbf{mIoU}(\%) & 79.75 & 83.07 & \textbf{85.25}  & 84.23 & 83.28 \\
\textbf{mBoundF}(\%) & 85.73 & 89.23 & \textbf{90.37} & 90.05 & 89.83\\
\end{tabular} }
\caption { Performance of Different Evolution Iterations}
\vspace{-0.15in}
\label{iterations}
\end{table}

\begin{table}[!h]
\centering
\scalebox{0.85} {
\begin{tabular}{l|ccccc}
\textbf{Number of Contour Points} & {32} & {64} & {128} & {256} \\ \hline
\textbf{mDice}(\%) & 75.76 & 85.56 & \textbf{93.76} & 91.26  \\
\textbf{mIoU}(\%) & 81.52 & 84.43 & \textbf{85.25}  & 83.42  \\
\textbf{mBoundF}(\%) & 80.73 & 86.25 & \textbf{90.37} & 90.15 \\
\end{tabular} }
\caption {Performance of Different Contour Point Numbers}
\label{number}
\vspace{-0.05in}
\end{table}

\subsubsection{\textbf{Parameter Settings}}

We further evaluate the effects of iteration count and contour point quantity on model performance using the MR\_AVBCE dataset. Optimal performance is observed at three iterations, as detailed in Table \ref{iterations}. Extending iterations beyond this threshold (e.g., to four or five) yields no additional benefits and may complicate training due to impaired gradient propagation, hindering network optimization. Additionally, we assess the influence of contour point numbers on segmentation accuracy, with findings presented in Table \ref{number}. The results reveal that 128 contour points achieve peak performance. A reduced count (e.g., 64) compromises the model’s ability to delineate complex organ boundaries, resulting in inferior segmentation outcomes. Conversely, increasing the number of points excessively (e.g., 256) significantly raises computational costs without yielding further performance gains. This is likely because additional points provide redundant information and increase the training difficulty. Thus, selecting 128 contour points strikes an optimal balance between precise boundary representation and computational efficiency.

\subsubsection{\textbf{Robustness to Contour Initialization}}
As a two-stage segmentation algorithm that integrates detection followed by contour evolution, the performance of the proposed method is potentially influenced by the accuracy of the initial contour localization. To evaluate the robustness of the algorithm to variations in initial bounding box placement, we conduct a sensitivity analysis using the Verse validation dataset. The initial bounding boxes are perturbed as follows:
\vspace{-0.2cm}
\begin{itemize}[leftmargin=*]
    \item \textbf{Positional Shift}: The center of the bounding box is displaced by 10\% and 20\% of the box's width and height.
    \item \textbf{Scale Jitter}: The box dimensions are scaled by 90\%/110\% and 80\%/120\% of their original size.
\end{itemize}
\vspace{-0.2cm}

The experimental results demonstrate robust performance, with a negligible decrease in segmentation accuracy (less than 0.5\% Dice coefficient reduction) for perturbations of $\pm$10\% and a modest reduction (approximately 2\% Dice coefficient) for perturbations of $\pm$20\%.

While the contour evolution process exhibits strong resilience to variations in the initial bounding box position, a failure in the detection phase inevitably precludes the initiation of the evolution process. To alleviate this problem, during training, contour evolution is initialized using ground-truth bounding boxes. This approach ensures stable learning and mitigates error propagation from the detection module, thereby enhancing the robustness of the contour evolution.

\vspace{-0.5em}
\section{Conclusion}

We present a new state-space-driven snake framework, Mamba Snake, for unified medical image segmentation, tackling the challenge of multi-scale structural heterogeneity. Mamba Snake designs multi-contour evolution as a hierarchical state space atlas, effectively modeling both macroscopic inter-organ topological relationships and microscopic contour refinements. A snake-specific state space module, Mamba Evolution Block (MEB), breaks the causal constraints of SSM, enabling efficient aggregation of temporal and spatial information. The energy shape priors and dual-classification synergy mechanism further improve evolution robustness and feature learning in heterogeneous data. Experimental results confirm Mamba Snake’s superiority over existing pixel-wise and contour-based methods, highlighting its potential as an effective clinical tool.

\bibliographystyle{ACM-Reference-Format}
\bibliography{sample-base}

\clearpage
\setcounter{page}{1}
\begin{center}
    \textbf{\Huge Supplementary Material}
\end{center}

\section{The Energy Shape Prior Map generation}
\label{sec:EnergyMapNetwork}
As shown in Table \ref{tab:efficientnetv2s_decoder}, a decoder module is added at the final stage of the network to adapt EfficientNetV2-S for energy map generation. This module consists of five upsampling stages, each including a \texttt{Conv2d} layer, \texttt{Batch Normalization}, \texttt{SiLU activation}, and a \texttt{ConvTranspose2d} layer with \texttt{output\_padding=1}. These stages progressively increase the spatial resolution, starting from \texttt{16×16} and upsampling to \texttt{512×512}. The final output layer consists of a \texttt{1×1 Conv2d} followed by a \texttt{Sigmoid activation}, which normalizes the values to the \texttt{[0,1]} range. An additional \texttt{normalization step} is applied after the \texttt{Sigmoid} function, scaling the output to \texttt{[0,255]}, as the energy map is inherently defined within this range. This ensures consistency with its intended representation and facilitates downstream processing.
\begin{table}[h]
    \centering
    \small 
    \setlength{\tabcolsep}{3pt} 
    \begin{tabular}{c|c|c|c|c}
        \hline
        Stage & Operator & Stride &  Channels & Layers \\
        \hline
        0  & Conv3x3 & 2  & 24   & 1  \\
        1  & Fused-MBConv1, k3x3 & 1  & 24   & 2  \\
        2  & Fused-MBConv4, k3x3 & 2  & 48   & 4  \\
        3  & Fused-MBConv4, k3x3 & 2  & 64   & 4  \\
        4  & MBConv4, k3x3, SE0.25 & 2  & 128  & 6  \\
        5  & MBConv6, k3x3, SE0.25 & 1  & 160  & 9  \\
        6  & MBConv6, k3x3, SE0.25 & 2  & 256  & 15 \\
        7  & Conv1x1 \& Pooling  & -  & 1280 & 1  \\
        \hline
        8  & Conv2d 3x3 + BatchNorm2d + SiLU & 1 & 512 & 1 \\
        9  & ConvTranspose2d 3x3 & 2  & 256  & 1  \\
        10 & Conv2d 3x3 + BatchNorm2d + SiLU & 1 & 128 & 1 \\
        11 & ConvTranspose2d 3x3 & 2  & 64   & 1  \\
        12 & Conv2d 3x3 + BatchNorm2d + SiLU & 1 & 32 & 1 \\
        13 & ConvTranspose2d 3x3 & 2  & 16   & 1  \\
        14 & Conv2d 3x3 + BatchNorm2d + SiLU & 1 & 8 & 1 \\
        15 & ConvTranspose2d 3x3 & 2  & 4    & 1  \\
        16 & Conv2d 3x3 + BatchNorm2d + SiLU & 1 & 2 & 1 \\
        17 & ConvTranspose2d 3x3 & 2  & 1    & 1  \\
        18 & Sigmoid Activation → Normalize (0-255) & -  & 1 & 1 \\
        \hline
    \end{tabular}
    \caption{EfficientNetV2-S with Decoder. The architecture removes the fully connected layers and uses transposed convolutions for upsampling, resulting in a final energy map with the same resolution as the input image.}
    \label{tab:efficientnetv2s_decoder}
\end{table}

\vspace{-1em}
\section{Detector Introduction}
The detector in Mamba Snake utilizes CenterNet \cite{centernet}, an anchor-free object detection algorithm that predicts the center point location of the object along with its associated attributes, such as width, height, and class, to achieve efficient object detection and classification. Unlike traditional anchor-based detection methods, CenterNet locates objects by generating a center point heatmap and then regresses the object's bounding box dimensions, simplifying and enhancing the accuracy of the detection process.

In Mamba Snake, CenterNet adopts DLA-34 \cite{DLA} as its backbone network. The output layer comprises three branches: heatmap, offset, and size, with corresponding output dimensions of \((W/R, H/R, C)\), \((W/R, H/R, 2)\), and \((W/R, H/R, 2)\), respectively, where \(R\) denotes the stride (set to 4 in this study) and \(C\) represents the number of organ classes. CenterNet identifies the center points of objects within the image, allowing the model to focus on target organs while mitigating interference caused by unclear boundaries and complex backgrounds. Additionally, the bounding box dimensions and aspect ratios provide rough morphological cues about the segmented objects, enabling the model to better adapt to organs with significant variations in shape and size.

It is worth noting that Mamba Snake only needs the detection boxes provided by the detector for initializing the polygonal contours. Therefore, the detector can be replaced by any other detection model, such as the Yolo V8 \cite{YOLOv8} or Mask R-CNN \cite{2017mask_r_cnn}.

\vspace{-1em}
\section{Contour Point Sampling}
\begin{figure}[!ht]
\centering
\setlength{\abovecaptionskip}{-0.03em}   
\includegraphics[width=0.48\textwidth]{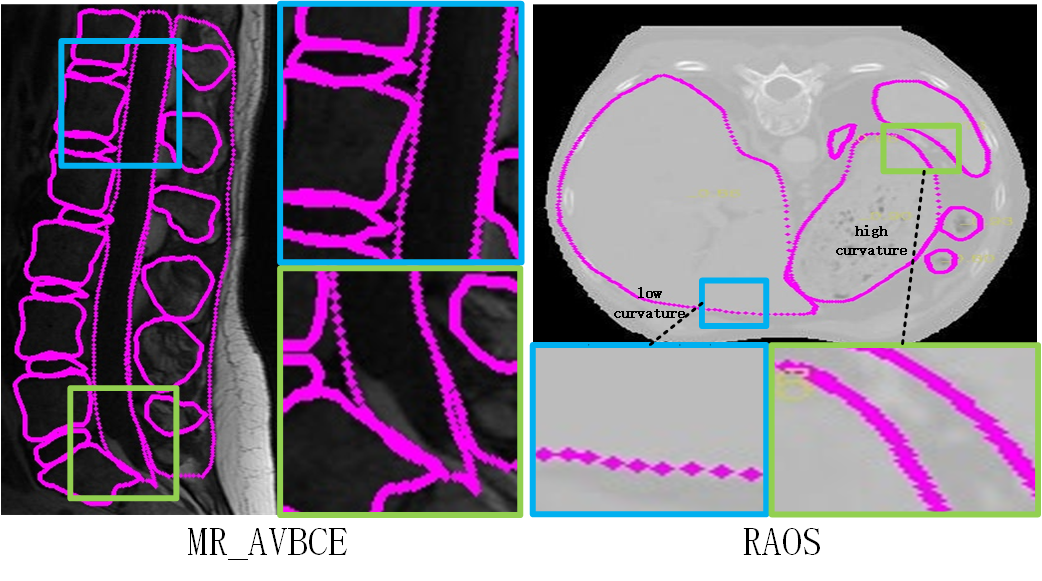} 
\caption{Visualization of contour points}
\label{fig5}
\vspace{-1.0em}
\end{figure}

As demonstrated in the point number experiment in the main text, the optimal performance is achieved when the number of contour points is set to 128. To better address the issue of insufficient concavity representation in regions with high curvature, we introduce a curvature penalty term ${L}_{curvature}$. The curvature penalty term is defined as follows:
\begin{equation}
\mathcal{L}_{curvature} = \sum_{i=1}^{N} \left( \kappa_i \cdot w(\kappa_i) \right)
\end{equation}
where \( \mathcal{L}_{curvature} \) is the total curvature penalty, \( \kappa_i \) is the curvature at the \( i \)-th contour point, and \( w(\kappa_i) \) is a weighting function \( w(\kappa_i) = \kappa_i^2 \) that increases with curvature. This encourages the Mamba Snake to place more contour points in areas with higher curvature, leading to more accurate contour representation. We visualized the 128 points for each contour on the MR\_AVBCE and RAOS datasets, and the results are presented as Figure \ref{fig5}.

\section{Data Processing}
Mamba Snake is designed to handle both 2D and 3D images by converting 3D volumes into a series of 2D slices for processing.
For instance, the VerSe dataset contains approximately 60 slices per CT scan, while the BTCV dataset includes 100 to 200 slices, and the RAOS dataset consists of around 200 slices per CT scan. In the VerSe dataset, we use sagittal slices for both training and evaluation, whereas for the BTCV and RAOS abdominal datasets, axial slices are utilized. All slices are uniformly cropped to a resolution of 512 × 512 pixels.

\begin{table*}[!h]
    \centering
    \small
    \renewcommand{\arraystretch}{1}
    \resizebox{\textwidth}{!}{
    \begin{tabular}{l|p{4cm}|p{5cm}|p{6cm}}
        \hline
        \textbf{Dataset} & \textbf{Modality and Samples} & \textbf{Annotations} & \textbf{Characteristics} \\
        \hline
        MR\_AVBCE & 600 MRI images from three institutions & 4601 individual vertebral segmentation masks & Includes approximately 820 tumor impacts, 120 degenerations, 20 artifacts, 270 blurred edges; significant differences in size, texture, and pathological deformation \\ 
        \hline
        VerSe & 374 CT scans from 355 patients & Segmentation masks for 256 individual vertebrae (C1 to L5, T13, L6) & Variations in size and shape across spinal regions \\ 
        \hline
        RAOS & 317 CT scans from two institutions & Manual segmentation masks for 17-19 abdominal organs & Covers cases from no organ resection to postoperative; diversity in organ size, shape, and anatomical changes \\ 
        \hline
        PanNuke & 2,655 image patches & Over 54000 cell nucleus segmentation masks & Variations in cell nucleus size and appearance across 19 tissue types and 5 cell categories \\ 
        \hline
        BTCV & 18 training and 12 testing abdominal CT scans & Segmentation masks for 8 abdominal organs per scan & Variability in organ size, shape, and complex anatomical relationships \\
        \hline
    \end{tabular}
    }
    \caption{Overview of various medical image segmentation datasets.}
    \label{tab:datasets}
    \vspace{-1.9em}
\end{table*}

\section{Dataset Introduction}
This study employs five influential segmentation datasets to assess the performance of Mamba Snake: MR\_AVBCE (spine, MRI) \cite{data_MRAVBCE}, VerSe (spine, CT) \cite{data_verse}, RAOS (abdomen, CT) \cite{data-RAOS}, BTCV (abdomen, CT) \cite{dataset_btcv}, and PanNuke (cells, microscopy) \cite{data-PanNuke}. These datasets cover a range of tissues and exhibit multi-scale structural heterogeneity. Detailed descriptions are provided below.

\textbf{MR\_AVBCE} \quad The MR\_AVBCE dataset \cite{data_MRAVBCE} poses substantial challenges for unified medical image segmentation. Specifically, it consists of 150 MRI images from the Affiliated Hangzhou First People’s Hospital, 407 from Qilu Hospital of Shandong University, and 43 from \textbf{S}aint Joseph Health Care Center of London, amounting to a total of 600 images. There is a significant variance in vertebrae sizes, textures, intensity distributions, and pathological morphology deformations among different patients. In total, the dataset contains 4601 vertebrae. Among them, approximately 820 vertebrae are affected by tumors, 120 by degenerative diseases, 20 by artifacts, and 270 by blurred edges due to low imaging quality. This subset, with pathological deformations and imaging issues, poses the greatest challenge as their training signals may be overshadowed by normal vertebrae.

\textbf{VerSe}\quad The VerSe dataset \cite{data_verse} is a substantial dataset for vertebra segmentation, consisting of 374 CT scans sourced from 355 patients. It includes voxel-level annotations for individual vertebrae across two subsets. The dataset covers 26 vertebrae, ranging from C1 to L5, along with the transitional vertebrae T13 and L6, annotated with labels 1 to 24. L6 and T13 are assigned labels 25 and 28, respectively. The mid-sagittal planes for each patient are used in our experiments.

\textbf{RAOS}\quad The RAOS dataset \cite{data-RAOS} is a clinical benchmark for abdominal organ segmentation. The patients represented in these images received treatments such as non-invasive therapy, surgery, radiation, and chemotherapy. The dataset is divided into Set-A (no organ resection), Set-B (surgery without organ removal), and Set-C (surgery with organ removal). Due to the smaller number of surgical cases, Set-A (comprising 317 CT scans in total) is used for network training and internal evaluation (220 for training and 67 for testing). The remaining 30 validation scans were not used because the hyperparameters were not optimized in our experiment. It also provides manual annotations for 17 organs in females and 19 in males, serving as a key resource for assessing model performance on complex abdominal cases. The mid-sagittal planes for each patient are used in our experiments.

\textbf{PanNuke} \quad The PanNuke dataset \cite{data-PanNuke} is utilized to evaluate the performance of our model in segmenting cell nuclei across a variety of tissue types. Due to time and resource constraints, we did not use the entire dataset for training and evaluation. Comprising over 54000 annotated nuclei spread across 2,655 images, each with a resolution of 256 \(\times\) 256 pixels, PanNuke is categorized into five different cell classes. These cell images were captured at 40\(\times\) magnification, with a pixel resolution of 0.25 \(\mu\)m/px, providing high-quality cellular details. This dataset serves as an essential resource for testing the cell segmentation capabilities of our model, offering both diverse tissue structures and challenging class distributions.

\textbf{BTCV} \quad The BTCV dataset \cite{dataset_btcv} is part of the Medical Segmentation Decathlon challenge. It consists of 18 training and 12 testing cases, each containing segmentations for 8 abdominal organs: (aorta, gallbladder, spleen, left kidney, right kidney, liver, pancreas, and stomach. This dataset presents challenges due to the variability in organ sizes, shapes, and their complex anatomical relationships, making it an ideal benchmark for evaluating the performance of segmentation models in handling multi-scale structural heterogeneity. The mid-sagittal planes for each patient are used in our experiments.

\vspace{-0.5em}
\section{Limitation}
We discusses several limitations of our contour-based segmentation approach under specific scenarios: 
(1) Handling Instances with Holes: While our contour segmentation method excels at delineating fine-grained boundary contours, it faces challenges when processing objects with internal holes. (2) Fine-Grained or Disconnected Structures: The contour-based model exhibits reduced performance when segmenting extremely small objects (e.g., objects spanning only a few pixels) or structures with topologically disconnected boundaries. In such scenarios, pixel-based segmentation methods may outperform our approach due to their ability to handle fragmented or minute structures. (3) Dependence on Detection: The success of the contour evolution process is contingent upon the initial detection of objects. If the detector fails to identify an object, the evolution phase cannot proceed, leading to inevitable segmentation failure. These limitations highlight areas for future optimization and improvement in contour-based segmentation.

\end{document}